\documentclass{article}

\usepackage{PRIMEarxiv}
\usepackage[utf8]{inputenc} 
\usepackage[T1]{fontenc}    
\usepackage{hyperref}       
\usepackage{url}            
\usepackage{booktabs}       
\usepackage{amsfonts}       
\usepackage{nicefrac}       
\usepackage{microtype}      
\usepackage{lipsum}
\usepackage{fancyhdr}       
\usepackage{graphicx}       
\graphicspath{{media/}}     
\usepackage{filecontents}
\usepackage{amsmath,amssymb,amsfonts}
\usepackage{algorithmic}
\usepackage{textcomp}
\usepackage{xcolor}
\usepackage[most]{tcolorbox}
\usepackage{framed}
\definecolor{shadecolor}{RGB}{0,200,230}
\usepackage{graphicx}
\usepackage{adjustbox}
\title{Translation Aligned Sentence Embeddings for Turkish Language
}

\author{
  Eren Unlu \\
  Datategy SAS \\
  Paris, France\\
  \texttt{eren.unlu@datategy.fr} \\
   \\
   \AND
   Unver Ciftci \\
   MATYZ Institute of Mathematics and Artificial Intelligence \\
   Tekirdag, Turkey \\
   \texttt{unver.ciftci@matyz.org} \\
}

\begin{document}
\maketitle

\begin{abstract}
Due to the limited availability of high quality datasets for training sentence embeddings in Turkish, we propose a training methodology and a regimen to develop a sentence embedding model. The central idea is simple but effective : is to fine-tune a pretrained encoder-decoder model in two consecutive stages, where the first stage involves aligning the embedding space with translation pairs. Thanks to this alignment, the prowess of the main model can be better projected onto the target language in a sentence embedding setting where it can be fine-tuned with high accuracy in short duration with limited target language dataset.
\end{abstract}


\section{Introduction}

With recent rapid advancements in Large Language Models (LLMs) and the following Retrieval Augmented Generation (RAG) applications, the importance of consistent and accurate sentence embeddings models have further increased \cite{siriwardhana2023improving}. One particular outcome of this interest in the market can be observed as the proliferation of open source and commercial initiatives to offer vector databases, which offer optimized service for the embeddings \cite{han2023comprehensive}\cite{pan2023survey}. Especially, in this landscape, the race to extend embeddings into longer contexts from short sentences has been getting increasingly competitive \cite{gunther2023jina}. So, it is reasonable to predict that generating semantically accurate and representative vector embeddings for various tasks will be at the heart of the evolving artificial intelligence ecosystem. 

Most of the sentence embedding models prefer to follow an architecture where first word or token embeddings are pooled, mostly by averaging. Then this pipeline is further fine-tuned in an end-to-end fashion to generate representative embeddings for general purpose or specific tasks. Though there are various approaches to curate supervised datasets from corpus (and/or human labeling/curation), usually the objective function is to minimize distances of embeddings of semantically closer phrases \cite{reimers2019sentence}\cite{wieting2015towards}.

Training plausible sentence embeddings models requires carefully curated datasets, as the definition of semantic proximity is generally vague and subjective, which makes the final dataset for fine-tuning as the primary source for the performance. Unfortunately, it is especially hard to find such datasets in relatively less represented languages such as Turkish \cite{fernando2023exploiting}\cite{eger2020probe}. Therefore, methods to automate dataset curation from high-resource languages and adapt the general purpose or sentence embedding models to low-resource languages are at the paramount of interest \cite{weeraprameshwara2022sinhala}. 

In this work, we propose a training methodology and regimen to fine-tune a \texttt{flan-t5-small} \cite{chung2022scaling}, which is a high performance language model trained for various instructions, in order to generate Turkish sentence embeddings. The model is accessible on huggingface hub with the name \texttt{myzens/turem512\_a}. Not only being trained on a large corpus and aligned for numerous human instructions, flan-t5 includes Turkish in its linguistic space. Even though Turkish prowess is almost non-existent for the small version, the fact that it pre-includes a Turkish tokenizer made it an ideal candidate in addition to its language understanding capabilities.  Also, having a pre-trained Turkish understanding up to a degree highlights the importance of it as a base model further, despite being insufficient.

The training strategy is composed of two distinct phases : First, the flan-t5 based sentence embedding pipeline with averaged token embeddings (flan-t5’s own token embeddings) is fine-tuned for a single simple sentence Turkish-English translation. Rather than retraining the base model for a neural machine translation (NMT) task, the idea is to train the sentence embedding model in a contrastive fashion to make embeddings of correct translation pairs closer in the latent space and incorrect ones farther. We refer to this step as the ``translation alignment'' phase. The main motivation of such an attempt is to pre-align the English embedding space of the base model which contains valuable information with Turkish. Though the idea of having a distinct pre-phase of sentence embedding fine-tuning for translation pairs is very simple, the results presented in this work indicate that it is highly effective. Several attempts exist to project translation pairs closer in the latent space, however such a formulation is the first in the literature to the best of our knowledge \cite{reimers2020making}\cite{yang2019multilingual}. Following the translation alignment phase, the sentence embedding model is further fine-tuned in Turkish with a regular pair of entailment sentences extracted from a machine translated supervised dataset. 

The performance of the proposed methodology is measured with another Turkish image annotation dataset, where cosine similarities between multiple labels describing the same images and random other ones are evaluated. Proposed structure and training regimen not only provides plausible Turkish representations but also decreases the training duration significantly.

\section{Related Work}

Vector representations of single sentences or longer textual units have a very central function in various tasks from RAG for LLM based applications to data mining. This key role of sentence embeddings has been increasingly highlighted in parallel with the evolution of the artificial intelligence and machine learning landscape. Early attempts for numerically representing sentences include simple bag-of-words methods and aggregation of word embeddings \cite{zhao2017fuzzy}. Later specialized neural architectures are developed to generate sentence embeddings such as \cite{kiros2015skip}\cite{conneau2018senteval}\cite{kashyap2023beyond}.

A remarkable turning point for sentence embeddings came with the advent of the language models like BERT \cite{devlin2018bert}. However, it is a well established finding that powerful language understanding models do not yield useful sentence representations when used directly \cite{kashyap2023beyond}\cite{ethayarajh2019contextual}.
For example, \cite{reimers2019sentence} proposed proper utilization and fine-tuning of BERT for sentence embeddings. Many other propositions exist such as post-processing language model embeddings \cite{li2020sentence} or adapting representations from multiple layers \cite{kim2021self}. The most widely adopted method currently is to fine-tune language models with specific labeled datasets or sentence pairings extracted from processed corpora. Most of these labeled datasets targeting sentence embeddings training or similar tasks are in English. It is particularly hard to find reliable datasets for low-resource languages. One intuitive solution is to use NMT to translate these datasets into target languages, which we also use a Turkish version in the second training phase \cite{budur2020data}. 

There exists various studies on developing sentence representations for low-resource languages by leveraging the linguistic prowess of a base model in high-resource languages.  Most intuitive one is to train language models or sentence embedding in a multi-linguistic fashion, which would inherently align the semantic representations of low-resource languages with the high-resource languages \cite{chaudhary2019low}. However, note that even low-resource language understanding of relatively large and well trained models on multilingual corpora like flan-t5 or BERT is very limited. \cite{artetxe2019massively} offers a single bidirectional (Long Short-Term Memory) LSTM encoder with shared byte-paired dictionary and an auxiliary decoder, jointly trained with multilingual corpora. \cite{feng2020language} proposes a dual encoder with a pre-trained BERT for bilingual sentence embedding learning with cross-lingual transfer through a translation ranking loss. Our work can be seen as a much simplified version of \cite{feng2020language} with a dual-stage fine-tuning. Authors in \cite{reimers2020making} present a knowledge distillation framework to create student models for target languages.

\section{Proposed Model, Datasets and Two-Stage Training}

A sentence embedding architecture based on the ``flan-t5-small'' model is proposed \cite{chung2022scaling}. Flan-t5 is particularly adept in language understanding as it is multi-instruction tuned. Popular word/token embedding pooling layer for initial stage is adopted. Flan-t5’s own tokenizer and embedder is used and mean aggregation is performed. As it includes Turkish in training corpora this allows tokenization in target language. In addition, despite lacking plausible performance in Turkish, the fact that it has been trained on this language shall help with the adaptability. 

The central idea of our work is elementary but proficient : training the sentence embedding model with pretrained flan-t5 weights for Turkish-English translation pairs. At the second stage, it is further trained with entailment pairs in Turkish. Theoretically, by initially making English-Turkish meanings closer in the embedding space, we can leverage the pretrained capacity of the base model by moving semantically similar sentences in target language. As the semantic sentence similarity in target language is performed in an additional stage, the initial aligned positions for English embeddings should be somehow properly positioned retrospectively. 

For the first stage, a simple single sentence English-Turkish translation pairs are used from the dataset in \cite{tiedemann2020tatoeba}. Multiple Negatives Ranking Loss (MNRL) is used for contrastive learning \cite{henderson2017efficient}. The first stage is trained for 120,000 batches with a batch size of 32, which roughly corresponds to 3.5 epochs. 2048 pairs are reserved initially for validating the results. Regular Adam optimizer is used with an initial learning rate of $10^{-5}$. 0.005 of weight decay is applied.

\begin{figure}[h]
\centering
\label{fig:fig_1}
\includegraphics[width=0.5\linewidth]{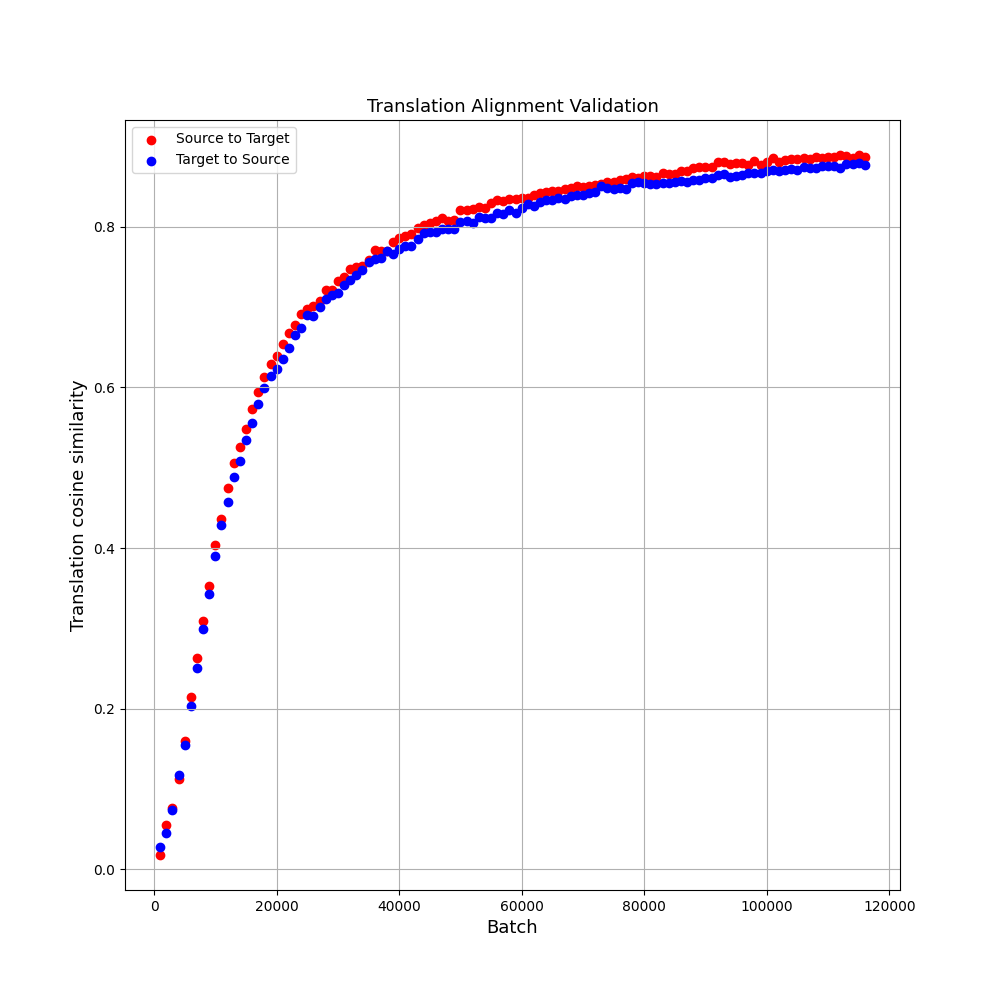}
\caption{First stage of translation pair training validation metric evolution (translation cosine similarities in both directions) on a separate test set of 2048 sentence pairs.}
\end{figure}

Second training phase is for contrastive training of similar Turkish sentences. Dataset offered by \cite{budur2020data} is used, which is a machine translated version of the original multi-genre NLI corpus dataset \cite{williams2017broad}. The training is designed as a regular similarity pairings including two sentences as in the first stage. For this purpose, only samples in the dataset for entailment are used. From a portion reserved for testing, 2000 validation pairings are generated; 1000 for proper matches and 1000 random pairings, where they are labeled for 1 and 0 cosine similarity, respectively. MNRL loss is used as the first translation alignment phase. Adam optimizer with an initial learning rate of $10^{-4}$ is used. A weight decay of 0.005 is applied. With a batch size of 16, the model is trained for approximately 16,000 batches which corresponds to roughly just 1.2 epochs. Pearson correlation of cosine similarities of validation samples are used as a measure to track the training performance.

\begin{figure}[h]
\centering
\label{fig:fig_2}
\includegraphics[width=0.5\linewidth]{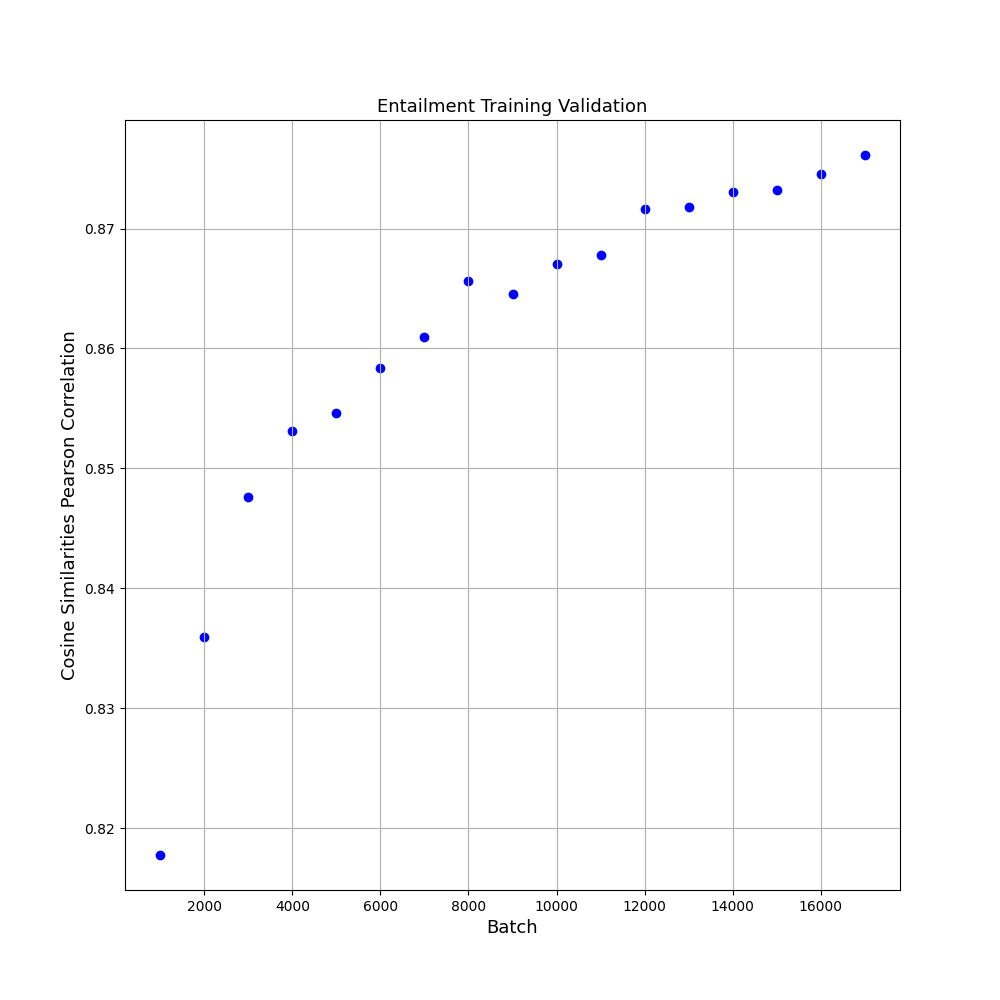}
\caption{Pearson correlation of cosine similarities of labeled 2000 samples is used as a validation metric to track training performance in the second phase.}
\end{figure}

Experimental Results

To evaluate the performance of the trained sentence embedding model we have used a separate dataset. Tasviret dataset is used which includes Turkish captions for images \cite{unal2016tasviret}. Certain images contain multiple different labels for the same images, therefore it allows us to measure similarity performance. Note that, not always necessarily, multiple annotations of the same image have similar meanings : Certain labels may describe totally different aspects of the same image. However, it still serves a valid benchmark. 8000 label pairs from the same images and 8000 label pairs from random images are used. Mean cosine similarity of sentence pairs from same images is 0.502, whereas random pairings yield 0.196. Note that, the actual performance can be expected to be much higher as not all image caption pairs are semantically similar. 

Several representative examples of cosine similarities for various image caption pairs are as follows : 
\begin{itemize}
  \item Sentence-1 : ``Siyah bir k\"{o}pek dalgalar{\i}n aras{\i}ndan \c{c}{\i}k{\i}yor.'' (A black dog emerges from the waves.)

  Sentence-2 : ``Siyah bir köpek dalgalar içinde koşmaya çalışıyor.'' (A black dog is trying to run in the waves.)

  Cosine Similarity : 0.862
   \item Sentence-1 : ``Donmu\c{s} bir zeminde elindeki alet ile bir delik a\c{c}maya \c{c}al{\i}\c{s}an bir adam ve yan{\i}nda duran k{\i}za\u{g}{\i}.'' (A man trying to make a hole with a tool in a frozen ground and his sled standing next to him.)

  Sentence-2 : ``Bal{\i}k avlamak i\c{c}in bir alet vas{\i}tas{\i}yla buzu delmeye \c{c}al{\i}\c{s}an bir adam.'' (A man trying to penetrate the ice with a tool to catch fish.)

  Cosine Similarity : 0.325

\item Sentence-1 : ``Deniz kenar{\i}nda s{\i}\u{g}l{\i}k bir yerde aerobik hareketleri yapan bir k{\i}z \c{c}ocu\u{g}u.'' (A girl doing aerobic exercises in a shallow place by the sea.)

  Sentence-2 : ``At kuyru\u{g}u sa\c{c}{\i} ile k\"{u}\c{c}\"{u}k k{\i}z denizde oynuyor.'' (Little girl with ponytail hair is playing in the sea.)

  Cosine Similarity : 0.391


\item Sentence-1 : ``Bir ku\c{s} \c{c}imlerin \"{u}zerinde ko\c{s}makta olan bir taz{\i} k\"{o}pe\u{g}inin pe\c{s}inden gidiyor.'' (A bird follows a greyhound running on the grass.)

  Sentence-2 : ``\c{C}imlerde ko\c{s}turan bir k\"{o}pek ve onun arkas{\i}ndan u\c{c}an bir ku\c{s}.'' (A dog running on the grass and a bird flying after it.)

  Cosine Similarity : 0.706


\item Sentence-1 : ``G\"{u}nbat{\i}m{\i}nda kar kaya\u{g}{\i} yapan biri.'' (Someone snowboarding at sunset.)

  Sentence-2 : ``\c{C}imlerde ko\c{s}turan bir k\"{o}pek ve onun arkas{\i}ndan u\c{c}an bir ku\c{s}.'' (Bir k\"{u}\c{c}\"{u}c\"{u}k k{\i}z \c{c}ocu\u{g}u plajda ko\c{s}arken.)

  Cosine Similarity : 0.038


\item Sentence-1 : ``Sahil kenar{\i}nda ko\c{s}an o\u{g}lan \c{c}ocuklar{\i}.'' (Boys running on the beach.)

  Sentence-2 : ``Dans eden \c{c}ocuklar.'' (Dancing children.)

  Cosine Similarity : 0.515


\item Sentence-1 : ``Bir kad{\i}n b\"{u}y\"{u}k bir \c{s}apka giyiyor.'' (A woman is wearing a large hat.)

  Sentence-2 : ``Bir k\"{o}pek ve bir inek \c{c}imler \"{u}zerinde.'' (A dog and a cow are on the grass.)

  Cosine Similarity : 0.205

\end{itemize}

We also visualized the Principal Component Analysis (PCA) reduced embeddings of Tasviret image captions. 10 random examples are chosen as it can be seen in Fig.3. These captions as illustrated in Fig. 3 are as follows :  
\begin{itemize}
\item 1 : ``Kaykay{\i} ile yerden baya y\"{u}kselmi\c{s} olan bir kaykayc{\i}.'' (A skateboarder who has risen high off the ground with her skateboard.)
\item 2 : ``Bir yer kaplamas{\i}n{\i} iki elinde ta\c{s}{\i}makta olan bir \c{c}al{\i}\c{s}an.'' (An employee carrying a floor mat in both hands.)
\item 3 : ``Tepenin \"{u}st\"{u}nden atlayan iki motokros\c{c}u.'' (Two motocrossers jumping over the hill.)
\item 4 : ``A\u{g}z{\i}nda tuttu\u{g}u renkli top ile \c{c}imlerin \"{u}zerinde ko\c{s}an kahverengi k\"{u}\c{c}\"{u}k bir k\"{o}pek.'' (A small brown dog running on the grass with a colorful ball in its mouth.)
\item 5 : ``Teleferikte oturan iki \c{c}ocuk.'' (Two children sitting on the cable car.)
\item 6 : ``Bir erkek \c{c}ocuk da\u{g}a kar\c{s}{\i} kar topu atacak.'' (A boy will throw a snowball against the mountain.)
\item 7 : ``Ba\c{s}{\i}n{\i} suya sokup \c{c}{\i}karm{\i}\c{s} bu s{\i}rada uzun sa\c{c}lar{\i}ndaki suyu d{\i}\c{s}ar{\i} s{\i}\c{c}ratan biri.'' (A person who dips her head into the water and splashes the water out of her long hair.)
\item 8 : ``\c{S}aka g\"{o}zl\"{u}\u{g}\"{u} takm{\i}\c{s} pembe ti\c{s}\"{o}rtl\"{u} bir k{\i}z \c{c}ocuk.'' (A girl in a pink t-shirt wearing joke glasses.)
\item 9 : ``Raftingte bottan azg{\i}n sulara d\"{u}\c{s}en macerac{\i}.'' (Adventurer falling from the boat into the raging waters during rafting.)
\item 10 : ``Bir siyah k\"{o}pek otlar{\i}n aras{\i}nda y\"{u}r\"{u}yor.'' (A black dog walks through the grass.)

\begin{figure}[h]
\centering
\label{fig:fig_3}
\includegraphics[width=0.5\linewidth]{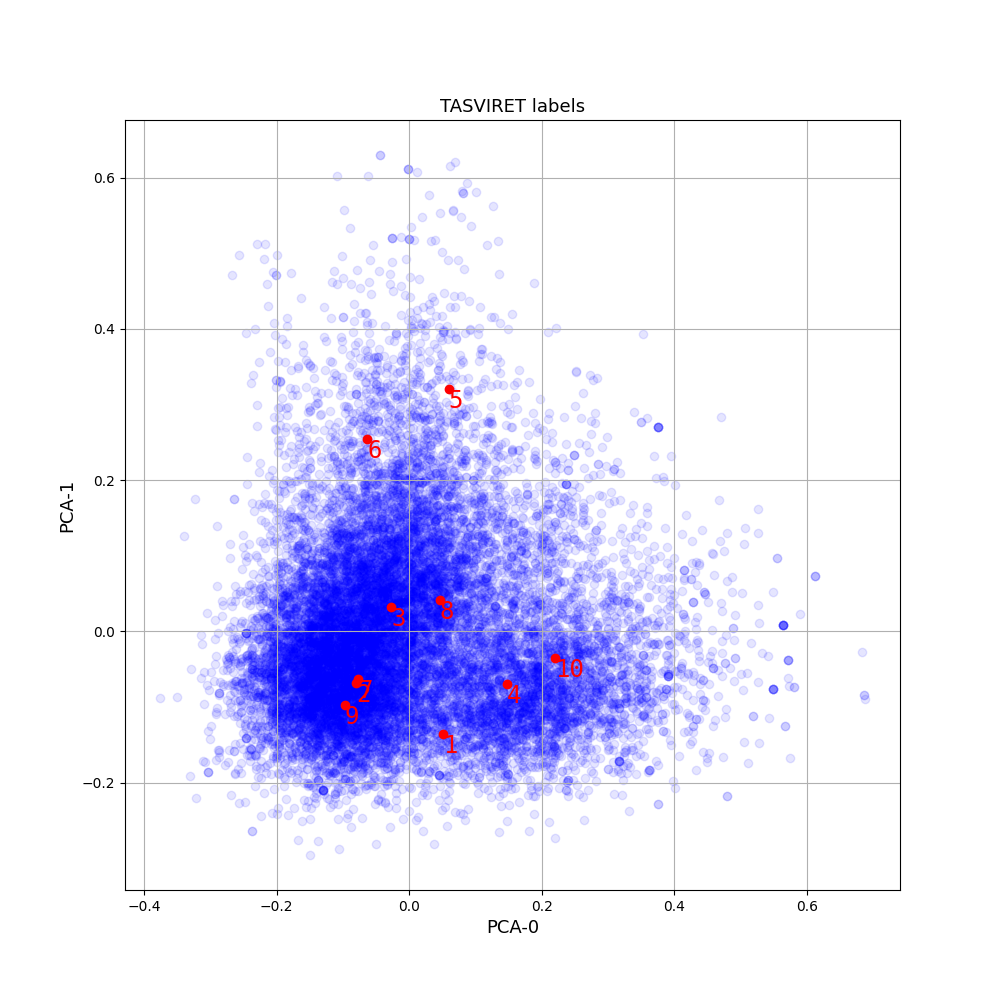}
\caption{PCA reduced embeddings of Tasviret image captions.}
\end{figure}
\end{itemize}

Note that, PCA illustration is for overall demonstrative purpose, where proximal disparities on the graph with real semantic similarities or dissimilarities are expected as the embedding space is highly reduced from 512 to 2 through a linear projection.

\section{Conclusion}
In this work, we have proposed a intuitive methodology and training regimen to develop a Turkish sentence embedding model based on ``flan-t5'' model. In order to overcome the scarcity of labeled datasets in low-resource languages, we fine-tune the base model based sentence embedding pipeline in two consecutive stages. At first stage, which we refer as translation alignment phase, the sentence embedder is trained for Turkish-English pairs. Next, the model is further fine-tuned with a Turkish sentence entailment pairs. This additive consecutive training strategy, theoretically allows to align semantics of Turkish phrases in the latent space by leveraging prowess of pre-trained base model in English. 


\end{document}